\documentclass[10pt,twocolumn,letterpaper]{article}

\usepackage{cvpr}              

\usepackage{graphicx}
\usepackage{amsmath}
\usepackage{amssymb}
\usepackage{booktabs}
\usepackage[table]{xcolor}

\usepackage[pagebackref,breaklinks,colorlinks]{hyperref}

\usepackage[capitalize]{cleveref}
\crefname{section}{Sec.}{Secs.}
\Crefname{section}{Section}{Sections}
\Crefname{table}{Table}{Tables}
\crefname{table}{Tab.}{Tabs.}
\crefname{figure}{Fig.}{Figs.}

\def\cvprPaperID{7606} 
\def\confName{CVPR}
\def\confYear{2022}

\begin{document}

\title{Do Explanations Explain? Model Knows Best}


\author{Ashkan Khakzar\textsuperscript{1}\thanks{denotes equal contribution}  , 
Pedram Khorsandi\textsuperscript{1}\textsuperscript{*}, 
Rozhin Nobahari\textsuperscript{2}\textsuperscript{*},
Nassir Navab\textsuperscript{1}\\
{\tt\small ashkan.khakzar@tum.de}\\
\\
\textsuperscript{1} Technical University of Munich, Germany \\
\textsuperscript{2} Mila, Quebec Artificial Intelligence Institute, Canada \\
}
\maketitle

\begin{abstract}
It is a mystery which input features contribute to a neural network's output.
Various explanation (feature attribution) methods are proposed in the literature to shed light on the problem.
One peculiar observation is that these explanations (attributions) point to different features as being important. 
The phenomenon raises the question, which explanation to trust?
We propose a framework for evaluating the explanations using the neural network model itself.
The framework leverages the network to generate input features that impose a particular behavior on the output.
Using the generated features, we devise controlled experimental setups to evaluate whether an explanation method conforms to an axiom.
Thus we propose an empirical framework for axiomatic evaluation of explanation methods.
We evaluate well-known and promising explanation solutions using the proposed framework.
The framework provides a toolset to reveal properties and drawbacks within existing and future explanation solutions.\footnote[1]{\url{https://github.com/CAMP-eXplain-AI/Do-Explanations-Explain}}

\end{abstract}

\section{Introduction}
\label{sec:intro}
Considering a neural network function, how do we know which features (patterns) within the input are important for its output? 
The problem is called feature attribution \cite{Lundberg2017,Sundararajan2017}, and the solutions are commonly known as explanation, attribution, or saliency methods.
There is an extensive list of explanation methods in the literature \cite{zhang2021fine,Lundberg2017,Sundararajan2017,Montavon2017,fong2017interpretable,fong2019understanding,Shrikumar2017,selvaraju2017grad,zhou2016learning,Schulz2020Restricting,Khakzar_2021_CVPR,khakzar2021explaining,Springenberg2015,srinivas2019full}.  
One peculiar observation is that these solutions point to different features as being important.
Though they are solutions to the same problem, feature attribution, the resulting explanations are curiously dissimilar.
The phenomenon raises the question, which explanation is correct?
Or are the explanations correct but revealing the problem in a different light?

One approach is to compare the explanations against ground truth (e.g., bounding box) annotations on the dataset \cite{selvaraju2017grad,zhou2016learning,zhang2018top}. But how do we know what is important for a human is also important for the model? There is no guarantee (or reason) that the model would use the same features as humans.
To resolve this issue, we need to take a step back and ask what it means for a feature to be “important” for an output. 
The intuitive approach is to remove the feature and observe the output behavior \cite{Samek2017,hooker2019benchmark,fong2019understanding}.
Such removal of evidence is indeed the foundation of many explanation approaches \cite{fong2017interpretable,fong2019understanding,Sundararajan2017,Lundberg2017,Schulz2020Restricting}.
However, such a conception could lead to ambiguities.
Consider the scenario of having equivalent features (e.g., repeated features), where the existence of each feature alone suffices for a specific output value. Add to the scenario that the removal of any of these features does not affect the output value.
In this case, the conception based on removal assigns zero importance to each feature.
However, a desirable property, in this case, could be assigning equal importance to each feature. 

The concept of importance can thus be further chiseled by specifying \emph{desirable properties} that an importance assignment method ought to satisfy.
Such \emph{desirable properties are formalized via axioms} \cite{Sundararajan2017,sundararajan2020many,Lundberg2017}.
The axiomatic view provides a complementary framework for evaluating feature attribution solutions.
Explanation methods can be evaluated whether they conform to an axiom.
The axiomatic view has the advantage that the methods can be mathematically proven to comply with a particular axiom.
For instance, solutions such as the Shapley value \cite{shapley1953value,Lundberg2017} and integrated gradients \cite{Sundararajan2017,sundararajan2020many} are proven to conform with particular axioms.
However, proofs can be broken in practical implementations. For instance, \cite{sundararajan2020many} show that inherent assumptions within methods that approximate the Shapley value result in methods not conforming with the axioms.
Moreover, certain conditions might be overlooked in proofs.
Thus \emph{experiments are required to test whether final solutions comply with the axioms}.
Even if methods are accompanied by elegant and solid mathematical derivations and proofs, they must comply with the axioms in observations in designed experiments. 
If they do not comply with axioms in experiments, we may revisit our assumptions and methodologies. Such is the way of the scientific method.

This work lays out an experimental framework for evaluating attribution solutions axiomatically.
We set up each experiment such that the solution can be tested whether it complies with a specific axiom. 
We generate input features that impose a particular behavior on the network’s input/output relationship.
Features are generated via optimization on the input space while the network parameters are kept constant.
Using optimization, we can impose the desired relationship between the generated input and the output. 
We can thus engineer setups to evaluate axioms.
For instance, one axiom that attribution methods are required to conform to is the Null-player axiom. 
The null-player axiom requires the following; If removal of a feature in all possible coalitions with other features does not affect the output, it should be assigned zero importance.
With our proposed framework, we can generate a null player feature for the neural network function. 
Subsequently, we can test different feature attributions solutions and check whether they assign importance to the null player feature. 
Thus we can test whether a solution conforms to the Null-player axiom. 
We also devise experiments to evaluate the explanations in terms of other desirable properties; The class-sensitivity and the feature-saturation.
With our framework, we evaluate well-known and recently introduced promising solutions.
With our experiments, we intend to reveal properties and drawbacks within existing explanations.

\section{Background and Related Work}
\subsection{Background}
We first introduce the feature attribution literature as our framework is designed to evaluate these methods. Then we introduce feature visualization/generation methods since they can be used within our framework.
\subsubsection{Explaining Predictions via Feature Attribution}

The feature attribution problem is concerned with identifying the input features that contribute to the output value. In neural networks literature, the problem is investigated from different perspectives. However, the underlying principle in many attribution solutions is that the notion of "importance" or contribution of a feature is associated with the effect of removal of that feature on the output value. The solutions can be roughly categorized as follows (some solutions belong to multiple categories).
\paragraph{Backpropagation} \cite{simonyan2013deep,baehrens2010explain} linearly approximate the network and propose the gradient as attribution. Deconvolution \cite{zeiler2014visualizing}, GuidedBackProp \cite{Springenberg2015} backpropagate a modified gradient.  Integrated Gradients \cite{Sundararajan2017} distributes the change in output with respect to a baseline input by integrating gradients between the two input states. LRP \cite{Montavon2017}, DeepLIFT\cite{Shrikumar2017} bacpropagate contribution layer-wise. The contribution notion in LRP and DeepLIFT is also grounded on \emph{removal}.
\paragraph{Perturbation/Removal} Methods in this category are explicitly grounded on the removal of features. They mask/perturb input features and observe the output change ~\cite{fong2017interpretable,fong2019understanding,qi2019visualizing, mishra2017local}. E.g., Extremal Perturbations\cite{fong2019understanding} searches for the smallest region in the input such that the keep the region preserves the target prediction. \cite{zeiler2014visualizing} propose occluding pixels or a patch of pixels and measure the output change. IBA \cite{Schulz2020Restricting} inserts an information bottleneck by removing hidden features (via replacing them with noise) and keeps the smallest region that preserves the predictive information. InputIBA \cite{zhang2021fine,zhang2021fine_arxiv} enables inserting the information bottleneck on the input.

\paragraph{Latent Features} CAM/GradCAM \cite{zhou2016learning,selvaraju2017grad} leverage activation values (aka network's attention) of convolutional layers. GradCAM++ uses different summation rules on layers and is applicable to all layers. IBA \cite{Schulz2020Restricting} also utilizes latent features. FullGrad leverages the activation, gradient and, bias values from all layers. PathwayGrad \cite{Khakzar_2021_CVPR} leverages critical pathways (pathway important neurons).

\paragraph{Game Theory} The attribution problem can be considered as credit assignment in cooperative game theory. This is achieved by presuming the network's function is a score function, and input features are players. A solution to this problem that satisfies several axioms is the Shapley Value. This notion is also grounded upon the removal of players and the effect of removal on score function. Shapley Value considers the removal of a player in all possible coalitions. Due to computational complexity, several approximations are proposed for neural networks. DeepSHAP \cite{Lundberg2017} backpropagates SHAP values via DeepLift \cite{Shrikumar2017} framework. It is recently shown \cite{sundararajan2020many} that Integrated Gradients \cite{Sundararajan2017} approximates Shapley value in continuous setting.

\subsubsection{Generating Features that Activate a Neuron}
The works in this category identify what input patterns/features activate a neuron and are commonly referred to as feature visualization.
In essence, the methods generate images that maximize certain neuron activations \cite{nguyen2016synthesizing,olah2017feature,erhan2009visualizing,simonyan2013deep,mahendran2015understanding,ulyanov2018deep}. This can be achieved by performing the optimization on the image while keeping networks parameters constant. However, due to the possibility of finding trivial solutions, the optimization is regularized in all these works. We can use any of these solutions within our framework. We opt for deep image prior \cite{ulyanov2018deep} (refer to \cref{sec:implement}). Another method that we utilize within our framework is the adversarial patch \cite{brown2017adversarial}.

\subsection{Related Work}
This section introduces the works that evaluate explanations. Our framework belongs to both "ground truth" and "axiomatic" categories. We discuss the differences to existing works in each section separately.
\paragraph{Do Explanations Explain?} An early work \cite{nie2018theoretical} demonstrates that Deconvolution \cite{zeiler2014visualizing} and Guided Backpropagation \cite{Springenberg2015} are reconstructing image features rather than explaining the prediction. Thus an explanation can be visually interpretable but not really be an explanation. The works in this section investigate whether an explanation method is indeed explaining the prediction and whether it can be trusted. Each work evaluates an explanation from a different vantage point. We categorize the works as follows:

\paragraph{Perturbation/Removal}
The objective of these works is to evaluate whether features identified as salient by attribution methods are indeed contributing to the output. The intuition behind them is that if the identified features are important, perturbing (removing) them changes the output relatively more. Sensitivity-N \cite{ancona2017towards} and \cite{Samek2017} use various perturbation schemes on the input and observe the output change. Remove-and-Retrain \cite{hooker2019benchmark} perturbs the input, then retrains the model and measures the accuracy drop. 

\paragraph{Ground Truth} These works compare the explanations with a ground truth of features that are important. Pointing game \cite{zhang2018top} and classic localization-based metrics \cite{selvaraju2017grad} use annotations on natural images done by humans. However, here there is an underlying assumption that the model is using the same features as humans, which is a crude assumption. To solve this issue, CLEVR XAI \cite{arras2020ground} proposes generating a synthetic dataset using CLEVR \cite{johnson2017clevr}. Thus we know what features are related to the output label, and we can use this as ground truth for evaluating explanations. The model is then trained on the generated dataset, and then explanations are then compared with the ground truth. The approach adds partial control to the experiment. However, there is no guarantee that the model picks up the intended features in the generated dataset. Our framework uses the model itself to generate the dataset. Therefore we can determine what features contribute or do not contribute to the model's output.

\paragraph{Axiomatic} Axiomatic approaches check whether the model complies with particularly desirable properties. Evaluation can be either theoretical, where the method is proven to satisfy an axiom (or a desirable property), \cite{Lundberg2017,Sundararajan2017,sundararajan2020many,sixt2019explanations}, or experimental. Sanity checks \cite{Adebayo2018} experimentally check whether randomizing network parameters change the explanation. Another desirable property is class sensitivity, i.e., the explanation should not be the same for different outputs (classes) if their contributing features differ. \cite{khakzar2020rethinking} provides a reasoning on why several methods are insensitive to parameter randomization and different classes. \cite{nie2018theoretical, Rebuffi2020} propose experiments to evaluate class-sensitivity on natural image datasets. However, on natural images, there is no guarantee that the model uses different features for different classes. Our framework provides a controlled setup as features are generated.

\section{Methodology}
The objective is to have a controlled experimental environment in which we control what features contribute or do not contribute to the output of the neural network function. In this environment, we can devise scenarios to test the explanations against axioms. To this end, we leverage the model itself and run an optimization on the input, thus controlling how features contribute to the output.

Importance or contribution is understood only with respect to a reference/baseline state ("removal" is setting feature values to a reference value). 
In our setup, we compute the contribution of a feature with respect to a reference of normal random noise. $X$ denotes the reference input. We refer to a group of pixels and their specific values as a "feature". In this work, we select a patch of pixels to form the feature. We denote the patch/feature by $f$ and a baseline input that has a feature $f$ added to it by $X_{\{f \}}$ and the neural network function for a target output $t$ by $\Phi_{t}(.)$. 
To generate the feature $f$ that corresponds to a target $t$ we generate a patch on the baseline input $X$ that activates the target $t$. Since the added feature changes the output value (by design), according to sensitivity axiom \cite{Sundararajan2017} it is guaranteed that it contributes to the output.
The optimization is performed only on patch $f$ (not on other areas of input $X$). The optimization loss for generating feature $f$ corresponding to target $t$ is denoted by $\mathcal{L}_{f}^{t}$. Depending on the scenario, $\mathcal{L}_{f}^{t}$ can be associated with either of the following. We can either generate a patch the maximizes the target value,
\begin{equation}\label{eq:feature_gen_max}
\begin{split}
    \min_{f} \; -\Phi_{t}(X_{\{f\}})
\end{split}
\end{equation}
or generate a patch that achieves a constant target value $c$, 
\begin{equation}\label{eq:feature_gen_const}
\begin{split}
    \min_{f} \; \mathcal{L}_{CE}(\Phi_{t}(X_{\{f\}}), c)
\end{split}
\end{equation}
where $\mathcal{L}_{CE}$ denotes cross-entropy loss.



\subsection{Null Feature}
The objective in this section is to devise a setup for testing the null feature axiom. A null feature is one that does not contribute to the output score. If a feature is a null feature, it is a desirable property for the explanation not to assign any contribution to that feature. based on cooperative game theory and attribution literature, null feature can be formally defined as follows.  Having a group of features (players), a feature is a null feature if its absence does not affect the output score function in all possible coalitions of features. I.e. if we have a set of $n$ features $\{f_{1}, ..., f_{n}\}$, a feature $f_{i}$ is null for output $\Phi_{t}(.)$ if $\Phi_{t}(X_{\{f_{i} \cup \mathcal{S}\}}) = \Phi_{t}(X_{\{\mathcal{S}\}})$, where $\mathcal{S}$ denotes all subsets of features excluding $f_{i}$, i.e. $\mathcal{S} \subset \{f_{1}, ..., f_{n}\} \setminus \{f_{i}\}$. Note that are $2^{N-1}$ possible coalitions.

In our experimental setup, we add two features to the baseline input $X$ (one can add more features and devise more complex or creative experiments). We add feature $f_{a}$ that corresponds to output $\Phi_{a}(.)$ and add another feature $f_{null}$ that corresponds to an output $\Phi_{b}(.)$ but is a null feature for output $\Phi_{a}(.)$. In order for the $f_{null}$ to be a null feature, its absence in all possible coalitions with $f_{a}$ should have no effect on output $\Phi_{a}(.)$. The are two possible coalitions, which are the subsets of ${f_{a}}$, namely ${f_{a}}$ and ${}$. Therefore, the output $\Phi_{a}(.)$ must stay constant when $f_{null}$ is removed when $f_{a}$ exists in baseline input $X$. The output $\Phi_{a}(.)$ must also stay constant when $f_{null}$ is added to the baseline $X$. Therefore, the optimization problem is defined as the following two concurrent optimizations,
\begin{equation}\label{eq:null_gen1}
    \min_{f_{a}} \; \mathcal{L}_{f_{a}}^{a}
\end{equation}
\begin{equation}\label{eq:null_gen2}
\begin{split}
    \min_{f_{null}} \; \mathcal{L}_{f_{null}}^{b} +  (\Phi_{a}(X_{\{f_{a},f_{null}\}}) -  \Phi_{a}(X_{\{f_{a}\}}) )^{2} \\ + (\Phi_{a}(X_{\{f_{null}\}}) -  \Phi_{a}(X_{\{\}}) ) ^{2}
\end{split}
\end{equation}
where $\mathcal{L}_{f_{a}}^{a}$ generates feature $f_{a}$ corresponding to output $\Phi_{a}(.)$. In \cref{eq:null_gen2} $\mathcal{L}_{f_{null}}^{b}$ generates a feature $f_{null}$ corresponding to output $\Phi_{b}(.)$. The second and third term in \cref{eq:null_gen2} try to make $f_{null}$ be a null feature for $\Phi_{a}(.)$ by removing it in possible coalitions with $f_{a}$. The result of the optimization is $X_{\{f_{a},f_{null}\}}$, which is a baseline noise image $X$ that contains patches/features $f_{a}$ and $f_{null}$. 

In this setup, we aim to test whether an explanation method attributes the output $\Phi_{a}(X_{\{f_{a},f_{null}\}})$ to the null feature. The proposed metric for evaluation is provided in \cref{sec:metric}

\subsection{Class Sensitivity}
Another property that is expected from an explanation method is class sensitivity, or in more general terms output sensitivity. Considering two outputs $\Phi_{a}(.)$ and $\Phi_{b}(.)$ of a neural network, if the contributing input features to the these outputs differ, the explanations for the outputs should also be different. To test such property we devise two scenarios:

\subsubsection{Single Feature Scenario}
In our first proposed setup, we only add one feature $f_{a}$ to the reference input $X$. The feature is generated such that it corresponds to the output $\Phi_{a}(.)$ but is a null feature for another output $\Phi_{a}(.)$. Therefore,
\begin{equation}
        \min_{f_{a}} \; \mathcal{L}_{f_{a}}^{a} + (\Phi_{b}(X_{\{f_{a}\}}) - \Phi_{b}(X_{\{ \}}))^{2}
\end{equation}
where the first term $\mathcal{L}_{f_{a}}^{a}$ generates the patch $f_{a}$ on reference input $X$, and the second term makes sure it is a null feature for output $\Phi_{b}(.)$. I.e. the removal of feature $f_{a}$ should not affect the output $\Phi_{b}(.)$. 

In this setup, the explanations for the two outputs $\Phi_{a}(.)$ and $\Phi_{b}(.)$ are compared. It is expected that the first explanation (for $\Phi_{a}(.)$) attributes the output (partly) to $f_{a}$. Whereas, the second expalanation (for $\Phi_{b}(.)$) should not attribute the prediction of $\Phi_{b}(.)$ to the the feature $f_{a}$. Our proposed metric for evaluating this effect is provided in \cref{sec:metric}.

\subsubsection{Double Feature Scenario}
In this setup we add two features $f_{a}$ and $f_{b}$ to the reference input $X$, each corresponding to the different outputs $\Phi_{a}(.)$ and $\Phi_{b}(.)$ respectively. In this setup the dominantly contributing feature to $\Phi_{a}(.)$ is feature $f_{a}$ and the dominantly contributing feature to $\Phi_{b}(.)$ is $f_{b}$.
Therefore we perform two concurrent optimizations. The first one,
\begin{equation}
        \min_{f_{a}} \; \mathcal{L}_{f_{a}}^{a} + (\Phi_{b}(X_{\{f_{a},f_{b}\}}) - \Phi_{b}(X_{\{f_{b}\}}))^{2}
\end{equation}
generates $f_{a}$ which contributes to output $\Phi_{a}(.)$ but its removal in the presence of feature $f_{b}$ does not affect output $\Phi_{b}(.)$. 
The second optimization,
\begin{equation}
      \min_{f_{b}} \; \mathcal{L}_{f_{b}}^{b} + (\Phi_{a}(X_{\{f_{a},f_{b}\}}) - \Phi_{a}(X_{\{f_{a}\}}))^{2}
\end{equation}
generates $f_{b}$ which contributes to output $\Phi_{b}(.)$ but its removal in the presence of feature $f_{a}$ does not affect output $\Phi_{a}(.)$. Thus the dominantly contributing feature for $\Phi_{a}(.)$ is $f_{a}$ and for $\Phi_{b}(.)$ is $f_{b}$. 

In this scenario we expect the explanations to switch from feature $f_{a}$ to $f_{b}$ when the output to be explained is changed from  $\Phi_{a}(.)$ to  $\Phi_{b}(.)$. Our proposed metric for capturing this metric is provided in section \cref{sec:metric}.

\subsection{Feature Saturation}
In this section, we devise a scenario where features saturate the output. Such that the features $f_{a1}$ and $f_{a2}$ together (i.e. $X_{\{f_{a1},f_{a2}\}}$) result in the same output value as when the features are added to reference input $X$ individually. To achieve this, we solve two optimizations concurrently,
\begin{equation}
        \min_{f_{a1}} \; \mathcal{L}_{f_{a1}}^{a} + (\Phi_{a}(X_{\{f_{a1},f_{a2}\}}) - \Phi_{a}(X_{\{f_{a2}\}}))^{2}
\end{equation}
where the first term generates $f_{a1}$ such that the output is equal to a constant value $c$. The second term makes sure that feature $f_{a1}$ removal from input does not affect the output when $f_{a2}$ is present. The second optimization does this procedure on the second feature $f_{a2}$,
\begin{equation}
        \min_{f_{a2}} \; \mathcal{L}_{f_{a2}}^{a} + (\Phi_{a}(X_{\{f_{a1},f_{a2}\}}) - \Phi_{a}(X_{\{f_{a1}\}}))^{2}
\end{equation}
In this setup, the existence of one of the features is sufficient for the prediction. As they contribute equally to the output, an explanation solution is expected to attribute the output equally to both features. Our proposed metric for evaluating this property is provided in \cref{sec:metric}.

\subsection{Metrics}\label{sec:metric}
In this section, we introduce our metrics for evaluating the properties in each of the generated setups. We denote an explanation generated for target output $\Phi_{t}(.)$ by $S_{t}$.
\paragraph{Null Feature Metric}
We define it as contributions assigned to null feature relative to total assigned contributions:
\begin{equation}
    \frac{\sum_{f_{a}}S_{a} }{\sum_{S_{a}}S_{a}}
\end{equation}
The sum operator $\sum_{f}S_{t}$ runs over all corresponding pixels in $S_t$ that are in patch $f$.

\paragraph{Class Sensitivity Metric}
In the Double Feature Scenario, we measure the class sensitivity by:
\begin{equation}
    \frac{\sum_{f_{a}\cup f_{b}}min(S_{a},\,S_{b})}{\sum_{f_{a}}S_{a} + \sum_{f_{b}}S_{b}}
\end{equation}
where the $min(S_{a},\,S_{b})$ is the pixel-wise minimum of $S_a$ and $S_b$.
In an extreme case, for the explanation method being indifferent towards the target class, the $min(S_{a},\,S_{b})$ would be equal to $S_a$ and $S_b$. Therefore, the metric evaluates to one. In the other extreme, where attributions shift from $f_{a}$ to $f_{b}$, the $min(S_{a},\,S_{b})$ and the metric is zero. 

For the Single Feature Scenario, the class sensitivity is determined through:
\begin{equation}
    corr(\frac{\overline{S_a-S_a\backslash f_a}}{\overline{S_a\backslash f_a}},\,\frac{\overline{S_b-S_b\backslash f_a}}{\overline{S_b\backslash f_a}})
\end{equation}
The term $\frac{\overline{S_t-S_t\backslash f}}{\overline{S_t\backslash f}}$ determines the average amount of contribution score inside patch $f$, devided by the average outside the patch. The higher correlation implies the method is attributing to the same feature for both outputs $\Phi_{a}$ and $\Phi_{b}$. 
\paragraph{Feature Saturation Metric}
To evaluate how the attribution is distributed between the features, we evaluate the correlation between attributions assigned to feature $f_{a1}$ and $f_{a1}$
\begin{equation}
    corr(\sum_{f_{a1}} S_{a},\,\sum_{f_{a2}}S_{a})
\end{equation}
A method that assigns the attribution to only one feature receives a lower score. 

\subsection{Implementation Details}\label{sec:implement}

\paragraph{Reference/Baseline Input}
Importance is understood with respect to a reference state. The reference is chosen such that it represents the missingness of features. In the vision domain, it is customary to use zero value \cite{Sundararajan2017,Shrikumar2017}, or noise \cite{Schulz2020Restricting}. In any case, our framework is not dependent on the reference. We do not make any assumptions about what features are in the reference. We ensure that a feature is null with respect to the reference, and our metric only considers the features and not the background.

\paragraph{Deep Prior Network} If we directly perform the optimizations on the patches without any regularization, it is easy to get trapped on local and trivial solutions. In this case, we may not achieve the target objective. Therefore regularizations are needed. In our work, we leverage "deep image prior" \cite{ulyanov2018deep} to limit the space of solutions and avoid trivial local solutions. Using deep image prior methodology, we add a decoder network with random weights and a random seed input behind the generated patch. In other words, the patch is parameterized by the prior network. The optimizations are thus done on the parameters of the prior network instead of the patch. 
In \cite{ulyanov2018deep} it is demonstrated that the untrained network does capture some of the low-level statistics of natural images. Therefore the generated patches also look interpretable to us. Visual interpretability is not required within our framework, though it can make the experiments more intuitive as the patches visually correspond to the target class.
%
%
\paragraph{Experimental Setup} 
The choice of the model does not affect the framework as long as the optimizations are solvable. 
The network is pre-trained on ImageNet \cite{deng2009imagenet}. However, the proposed framework does not depend on the network being trained. For a random network the generated features would not "look" interpretable.

\setlength{\fboxsep}{0pt}%

\begin{figure*}[!ht]
\centering
    \begin{subfigure}[b]{.1\textwidth}
      \centering
      \caption*{Image}
      \fbox{\includegraphics[width=.95\linewidth]{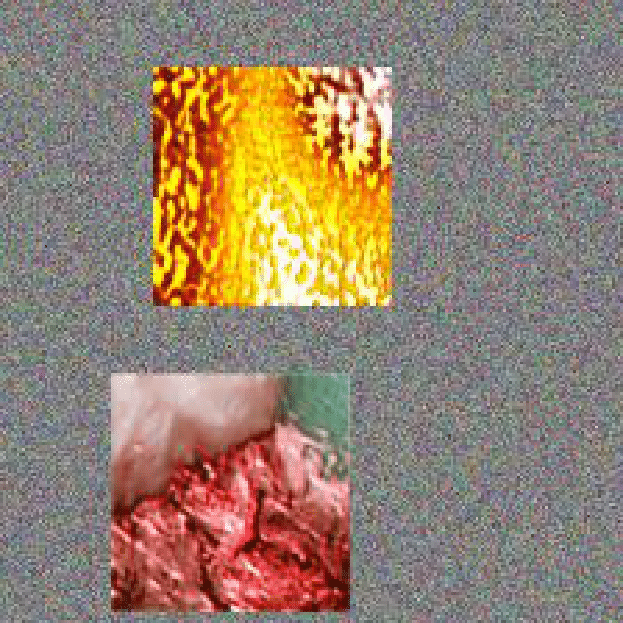}}
    \end{subfigure}%
    \begin{subfigure}[b]{.1\textwidth}
      \centering
      \caption*{\centering\scriptsize{GradCAM}}
      \fbox{\includegraphics[width=.95\linewidth]{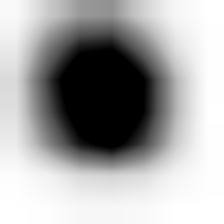}}
    \end{subfigure}%
    \begin{subfigure}[b]{.1\textwidth}
      \centering
      \caption*{\centering\scriptsize{GradCAM++}}
      \fbox{\includegraphics[width=.95\linewidth]{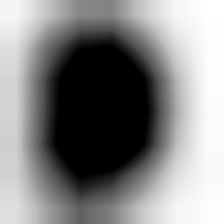}}
    \end{subfigure}%
    \begin{subfigure}[b]{.1\textwidth}
      \centering
      \caption*{\centering\scriptsize{Gradient}}
      \fbox{\includegraphics[width=.95\linewidth]{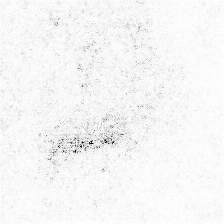}}
    \end{subfigure}%
    \begin{subfigure}[b]{.1\textwidth}
      \centering
      \caption*{\centering\scriptsize{FullGrad}}
      \fbox{\includegraphics[width=.95\linewidth]{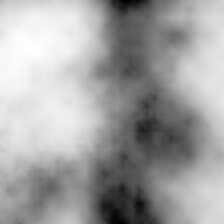}}
    \end{subfigure}%
    \begin{subfigure}[b]{.1\textwidth}
      \centering
      \caption*{\centering\scriptsize{GuidedBackProp}}
      \fbox{\includegraphics[width=.95\linewidth]{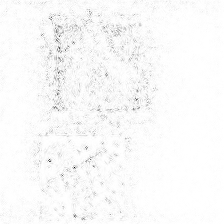}}
    \end{subfigure}%
    \begin{subfigure}[b]{.1\textwidth}
      \centering
      \caption*{\centering\scriptsize{Integrated Gradients}}
      \fbox{\includegraphics[width=.95\linewidth]{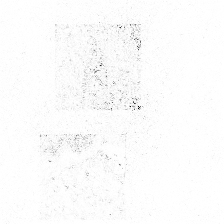}}
    \end{subfigure}%
    \begin{subfigure}[b]{.1\textwidth}
      \centering
      \caption*{\centering\scriptsize{DeepSHAP}}
      \fbox{\includegraphics[width=.95\linewidth]{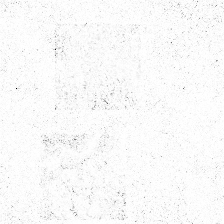}}
    \end{subfigure}%
    \begin{subfigure}[b]{.1\textwidth}
      \centering
      \caption*{\centering\scriptsize{IBA}}
      \fbox{\includegraphics[width=.95\linewidth]{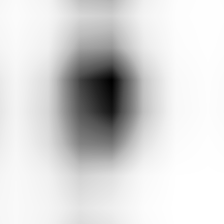}}
    \end{subfigure}%
    \begin{subfigure}[b]{.1\textwidth}
      \centering
      \caption*{\centering\scriptsize{Extermal Perturbation}}
      \fbox{\includegraphics[width=.95\linewidth]{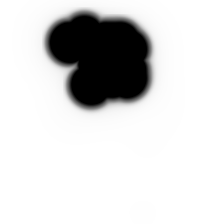}}
    \end{subfigure}%
  \caption{\textbf{Null Feature Experiment}: The image on the left represents the generated features on the reference (noise) input. The features are generated using the model itself. Within the image, the lower feature (patch) is generated such that it is a null feature for the output. The rest of the images represent different explanations. As the second feature is a null feature, an explanation method should not assign importance to it. We observe that GradCAM, IBA, and Extremal Perturbation perform best in avoiding the null feature.}
 \label{fig:null_player}
\end{figure*}
\setlength{\fboxsep}{0pt}%

\captionsetup[subfigure]{labelformat=empty}

\begin{table*}[t]
    \centering
    \rowcolors{5}{}{gray!10}
    \begin{tabular}{*5c}
        \toprule
        & & \multicolumn{2}{c}{Class Sensitivity} & \\
        \cmidrule(lr){3-4}
        Method & Null & Double Feature  & Single Feature & Feature \\    
           & Feature & Scenario & Scenario     & Saturation     \\
        \midrule
        GradCAM & 0.135 & 0.176 & 0.050 & 0.243\\
        GradCAM++ & 0.452 & 0.469 & 0.845 & -0.571\\
        Gradient & 0.835 & 0.469 & 0.684 & 0.310\\
        FullGrad & 1.00 & 0.931 & 0.951 & -0.130\\
        GuidedBackProp & 0.704 & 0.555 & 0.979 & 0.703\\
        IntegretedGradient & 0.534 & 0.344 & 0.759 & 0.212\\
        DeepSHAP & 1.03 & 0.507 & 0.934 & 0.221\\
        IBA & 0.211 & 0.191 & 0.295 & -0.223\\
        ExtermalPerturbation & 0.047 & 0.039 & 0.759 & -0.680\\
        \bottomrule
    \end{tabular}
    \caption{Evaluation of Explanations with the Framework: \textbf{1) Null Feature}: Null feature experiment evaluates the extent to which each explanation attributes the output to a null feature. In this metric, the less the value, the better. Extremal Perturbation \cite{}, GradCAM, and IBA are the favorable methods from this null feature perspective. \textbf{2) Class Sensitivity}: For both experiments, the lower the value, the better 1) Double Feature Scenario: In the case where two features corresponding to two different classes are present, Extremal Perturbation, IBA, and GradCAM attribute to the correct feature when applied to the two outputs. 2) Single Feature Scenario: In the case where only feature is present, explanations for two different outputs are similar to all methods except GradCAM and, to some extent, IBA. \textbf{3) Feature Saturation}: the experiment evaluates how explanations distribute the importance between saturated features. In this metric, the higher the value, the better. The notable observation is Extremal Perturbation, as it identifies only one of the features as important.}
    \label{tab:experiments}
\end{table*}

\paragraph{Patch Optimization}
During optimization steps, we place the patch in different locations to ensure that the results do not depend on the patch's location.
\paragraph{Focal Loss} The optimization problems in the framework have multiple losses. Therefore it is a challenge to balance the optimization between the losses. We use the focal loss \cite{guo2018dynamic} method to balance the optimization.

\section{Results and Discussion}

The objective of our proposed framework is to reveal insights and shortcomings regarding explanation methods. We evaluate various explanation methods from different categories. The methods are chosen based on their prevalence in the literature. DeepSHAP and IntegratedGradient are theoretically axiomatic methods. GradCAM and GradCAM++ are two popular methods that leverage network attention. We also evaluate the recently introduced FullGrad from this family. In addition, we evaluate two recent promising solutions, IBA and Extremal Perturbations.

\begin{table*}[ht]
    \centering
    \rowcolors{5}{}{gray!10}
    \begin{tabular}{*9c}
        \toprule
        &\multicolumn{4}{c}{IBA} & \multicolumn{4}{c}{GradCAM++} \\
        \cmidrule(lr){2-5}\cmidrule(lr){6-9}
        Metric & layer 1 & layer 2 & layer 3 & layer 4 & layer 1 & layer 2 & layer 3 & layer 4\\
        \midrule
        Null Feature & 0.315 & 0.311 & 0.201 & 0.211 & 0.827 & 0.906 & 0.815 & 0.453 \\
        Double Feature Scenario & 0.327 & 0.337 & 0.207 & 0.191 & 0.977 & 0.948 & 0.899 & 0.469 \\
        Single Feature Scenario & 0.219 & 0.237 & 0.158 & 0.295 & 0.979 & 0.823 & 0.761 & 0.845 \\
        \bottomrule
    \end{tabular}
    \caption{Evaluations of IBA and GradCAM++ explanations for various layers of ResNet:
IBA and GradCAM++ are applicable to different layers of convolutional networks. However, we observe that as we move towards earlier layers (toward the input), more attribution is assigned to the null feature. We also observe the same trend with class sensitivity. The results significantly deteriorate for GradCAM++ (in both experiments, the lower the value, the better). It is thus advisable to apply these explanations to deep layers.}
    \label{tab:multilayer}
\end{table*}
\subsection{Null Feature}

The null feature experiment checks whether an explanation attributes the output to a null feature. I.e., it checks whether the explanation method identifies the null feature as important. The framework guarantees that the null feature is not contributing. Using the framework, we generate 1000 inputs. For each input sample, the feature is generated for a random output.
We then proceed and compute the null feature metric on each generated input and report the average in \cref{tab:experiments}. 
An example generated input is in \cref{fig:null_player}. 

The results show that many explanation methods are assigning contribution to the null feature.
FullGrad, DeepSHAP, Gradient, and GuidedBackProp perform the worst in this experiment. This performance may point to the fact that these methods identify all features within the input as important. It is previously shown \cite{} that GuidedBackProp reconstructs image features rather than explaining the prediction, and our results are aligned with the finding. It can also be inferred that gradient is also sensitive to all features in the input. DeepSHAP is widely known as a solid method as it involves SHAP. However, it also has a backpropagation mechanism (as it is engineered on DeepLift). It seems the backpropagation is the culprit, as other gradient methods also fail this experiment. FullGrad does a weighted sum of gradients and biases of all layers. The gradients in the early layers can be the culprit in this case.

We observe that GradCAM rarely assigns attribution to the null feature. And the assigned values may be due to CAM's low resolution. GradCAM uses gradients/activations of deep layers and is not comparable to the aforementioned methods. IBA and extremal perturbation are both grounded on the removal of features. We see that they also avoid attributing to the null feature.

\begin{figure}[ht]
\centering
    \begin{subfigure}{\linewidth}
    \caption{GradCAM++}
      \centering
      \begin{subfigure}{.25\linewidth}
      \centering
      \fbox{\includegraphics[width=.95\linewidth]{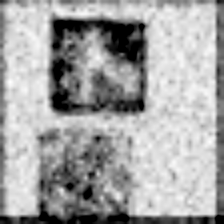}}
      \caption{\centering\scriptsize{layer 1}}
    \end{subfigure}%
      \begin{subfigure}{.25\linewidth}
      \centering
      \fbox{\includegraphics[width=.95\linewidth]{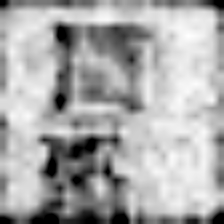}}
      \caption{\centering\scriptsize{layer 2}}
    \end{subfigure}%
      \begin{subfigure}{.25\linewidth}
      \centering
      \fbox{\includegraphics[width=.95\linewidth]{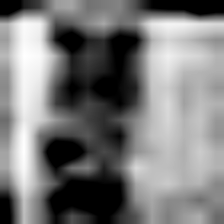}}
      \caption{\centering\scriptsize{layer 3}}
    \end{subfigure}%
      \begin{subfigure}{.25\linewidth}
      \centering
      \fbox{\includegraphics[width=.95\linewidth]{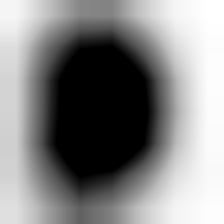}}
      \caption{\centering\scriptsize{layer 4}}
    \end{subfigure}%
    \end{subfigure}%
    \\    
    \begin{subfigure}{\linewidth}
    \caption{IBA}
      \centering
      \begin{subfigure}{.25\linewidth}
      \centering
      \fbox{\includegraphics[width=.95\linewidth]{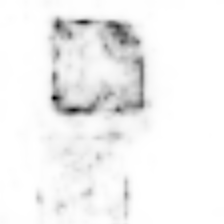}}
      \caption{\centering\scriptsize{layer 1}}
    \end{subfigure}%
      \begin{subfigure}{.25\linewidth}
      \centering
      \fbox{\includegraphics[width=.95\linewidth]{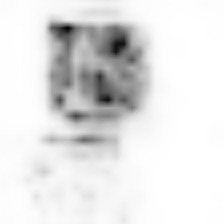}}
      \caption{\centering\scriptsize{layer 2}}
    \end{subfigure}%
      \begin{subfigure}{.25\linewidth}
      \centering
      \fbox{\includegraphics[width=.95\linewidth]{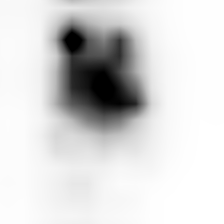}}
      \caption{\centering\scriptsize{layer 3}}
    \end{subfigure}%
      \begin{subfigure}{.25\linewidth}
      \centering
      \fbox{\includegraphics[width=.95\linewidth]{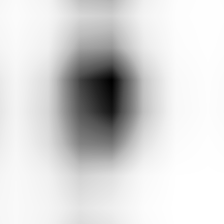}}
      \caption{\centering\scriptsize{layer 4}}
    \end{subfigure}%
    \end{subfigure}%
  \caption{Null Feature Experiment for IBA and GradCAM++ on different layers of a network. The second (lower) feature is a null feature. We observe that as we move toward earlier layers, the explanations attribute to the null feature for both methods.}
 \label{fig:null_player:multi_layers}
 
\end{figure}

We also evaluate IBA and GradCAM++ on different layers of a ResNet network. Among the advantages of these methods is that they can be applied to early layers to produce higher resolution maps. However, we observe in \cref{fig:null_player:multi_layers} and \cref{tab:multilayer} that as we move towards early layers, the methods attribute to the null feature.

\subsection{Class Sensitivity}
\subsubsection{Double Feature Scenario}
\begin{figure*}[ht]
\centering
    \begin{subfigure}[b]{.1\textwidth}
      \centering
      \caption{\centering\scriptsize{Image}}
      {\fbox{\includegraphics[width=.95\linewidth]{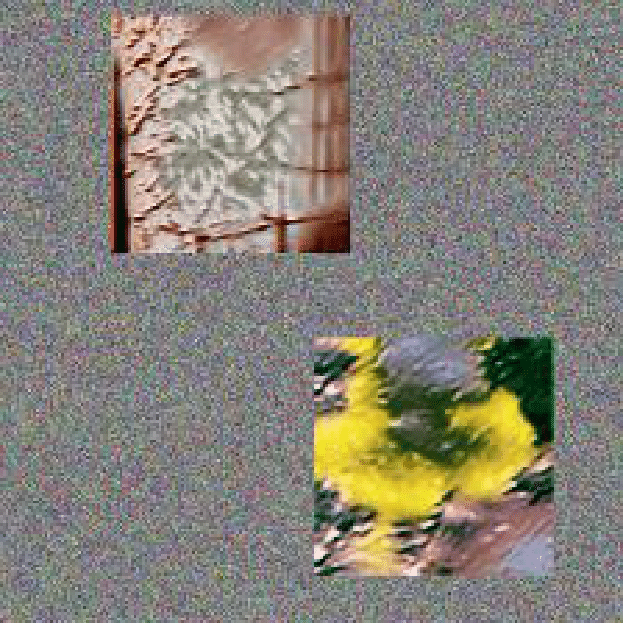}}}
    \end{subfigure}%
    \begin{subfigure}[b]{.1\textwidth}
      \centering
      \caption{\centering\scriptsize{GradCAM}}
      \fbox{\includegraphics[width=.95\linewidth]{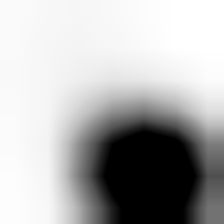}}
      \fbox{\includegraphics[width=.95\linewidth]{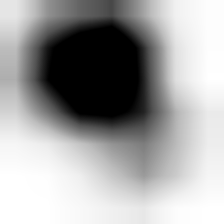}}
    \end{subfigure}%
    \begin{subfigure}[b]{.1\textwidth}
      \centering
      \caption{\centering\scriptsize{GradCAM++}}
      \fbox{\includegraphics[width=.95\linewidth]{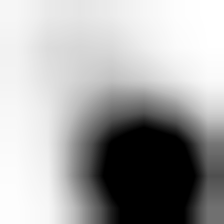}}
      \fbox{\includegraphics[width=.95\linewidth]{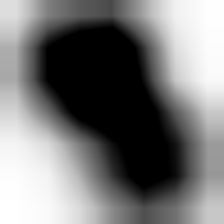}}
    \end{subfigure}%
    \begin{subfigure}[b]{.1\textwidth}
      \centering
      \caption{\centering\scriptsize{Gradient}}
      \fbox{\includegraphics[width=.95\linewidth]{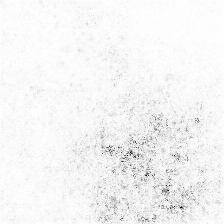}}
      \fbox{\includegraphics[width=.95\linewidth]{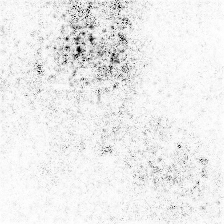}}
    \end{subfigure}%
    \begin{subfigure}[b]{.1\textwidth}
      \centering
      \caption{\centering\scriptsize{FullGrad}}
      \fbox{\includegraphics[width=.95\linewidth]{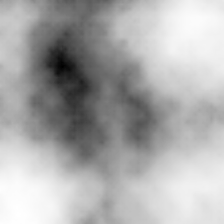}}
      \fbox{\includegraphics[width=.95\linewidth]{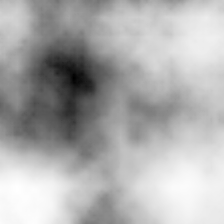}}
    \end{subfigure}%
    \begin{subfigure}[b]{.1\textwidth}
      \centering
      \caption{\centering\scriptsize{GuidedBackProp}}
      \fbox{\includegraphics[width=.95\linewidth]{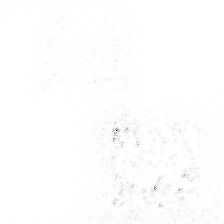}}
      \fbox{\includegraphics[width=.95\linewidth]{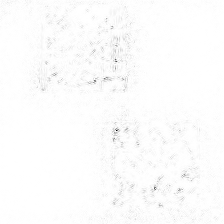}}
    \end{subfigure}%
    \begin{subfigure}[b]{.1\textwidth}
      \centering
      \caption{\centering\scriptsize{Integrated Gradients}}
      \fbox{\includegraphics[width=.95\linewidth]{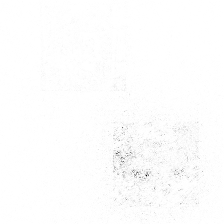}}
      \fbox{\includegraphics[width=.95\linewidth]{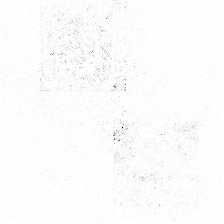}}
    \end{subfigure}%
    \begin{subfigure}[b]{.1\textwidth}
      \centering
      \caption{\centering\scriptsize{DeepSHAP}}
      \fbox{\includegraphics[width=.95\linewidth]{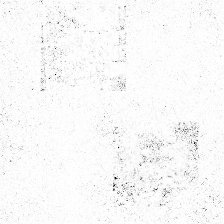}}
      \fbox{\includegraphics[width=.95\linewidth]{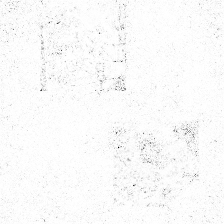}}
    \end{subfigure}%
    \begin{subfigure}[b]{.1\textwidth}
      \centering
      \caption{\centering\scriptsize{IBA}}
      \fbox{\includegraphics[width=.95\linewidth]{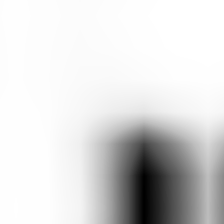}}
      \fbox{\includegraphics[width=.95\linewidth]{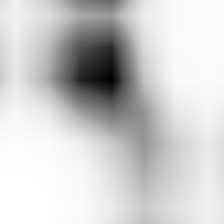}}
    \end{subfigure}%
    \begin{subfigure}[b]{.1\textwidth}
      \centering
      \caption{\centering\scriptsize{Extremal Perturbation}}
      \fbox{\includegraphics[width=.95\linewidth]{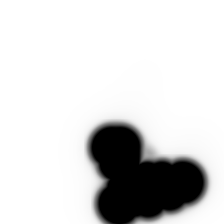}}
      \fbox{\includegraphics[width=.95\linewidth]{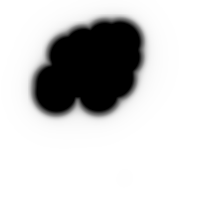}}
    \end{subfigure}%
  \caption{\textbf{Class Sensitivity - Double Feature Scenario}: The image on the left represents the generated features on the reference input. The features are generated such that each corresponds to a different output. The lower feature (patch) corresponds to the first output (first row), and the other patch corresponds to the second output (second row). It is expected that explanations for the two outputs differ and attribute to the corresponding feature of each output. GradCAM, IBA, and Extremal Perturbation manifest this property.}
 \label{fig:class_sensitivity}
\end{figure*}

The objective is to observe how the explanations for two different outputs differ when both outputs have corresponding features present. The metric results are presented in \cref{tab:experiments}, and visual examples are presented in \cref{fig:class_sensitivity}.
We observe that GradCAM, IBA, and Extremal Perturbation attribute the corresponding features when explaining the different outputs. FullGrad produces the same explanation when applied to the two outputs. We observe that Gradient, GuidedBackProp, DeepSHAP, and IntegratedGradient slightly switch to the other feature when the other output is explained.
We also perform layerwise experiments for IBA and GradCAM++. We observe that the explanations become less class sensitive in earlier layers.

\subsubsection{Single Feature Scenario}
\begin{figure*}[ht]
\centering
    \begin{subfigure}[b]{.1\textwidth}
      \centering
      \caption{\centering\scriptsize{Image}}
      {\fbox{\includegraphics[width=.95\linewidth]{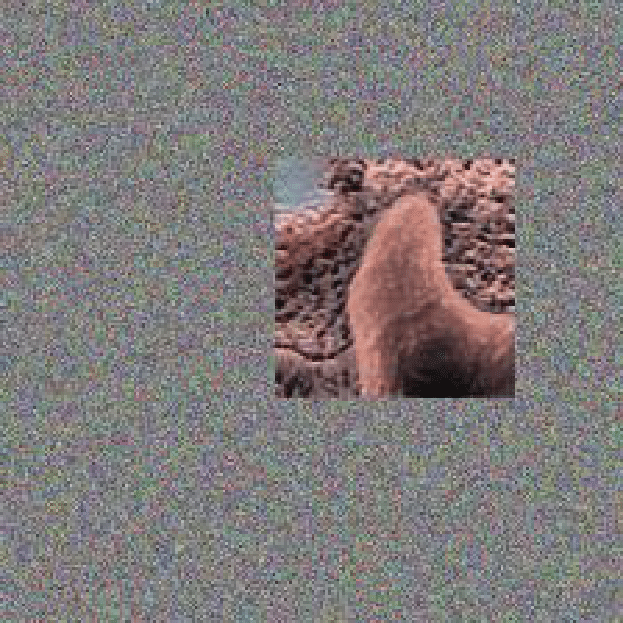}}}
    \end{subfigure}%
    \begin{subfigure}[b]{.1\textwidth}
      \centering
      \caption{\centering\scriptsize{GradCAM}}
      \fbox{\includegraphics[width=.95\linewidth]{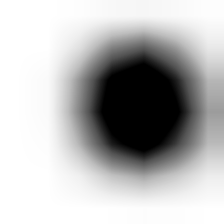}}
      \fbox{\includegraphics[width=.95\linewidth]{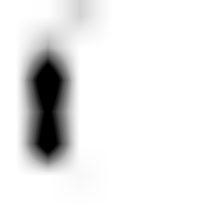}}
    \end{subfigure}%
    \begin{subfigure}[b]{.1\textwidth}
      \centering
      \caption{\centering\scriptsize{GradCAM++}}
      \fbox{\includegraphics[width=.95\linewidth]{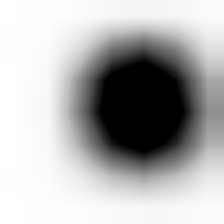}}
      \fbox{\includegraphics[width=.95\linewidth]{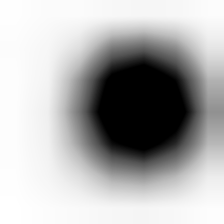}}
    \end{subfigure}%
    \begin{subfigure}[b]{.1\textwidth}
      \centering
      \caption{\centering\scriptsize{Gradient}}
      \fbox{\includegraphics[width=.95\linewidth]{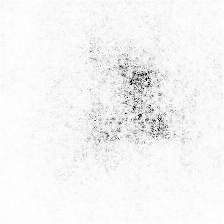}}
      \fbox{\includegraphics[width=.95\linewidth]{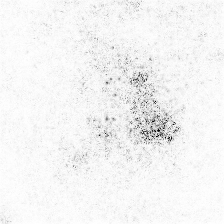}}
    \end{subfigure}%
    \begin{subfigure}[b]{.1\textwidth}
      \centering
      \caption{\centering\scriptsize{FullGrad}}
      \fbox{\includegraphics[width=.95\linewidth]{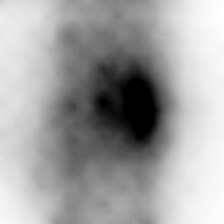}}
      \fbox{\includegraphics[width=.95\linewidth]{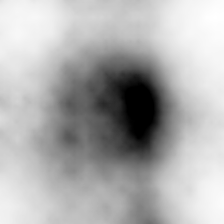}}
    \end{subfigure}%
    \begin{subfigure}[b]{.1\textwidth}
      \centering
      \caption{\centering\scriptsize{GuidedBackProp}}
      \fbox{\includegraphics[width=.95\linewidth]{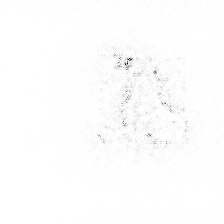}}
      \fbox{\includegraphics[width=.95\linewidth]{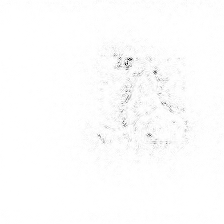}}
    \end{subfigure}%
    \begin{subfigure}[b]{.1\textwidth}
      \centering
      \caption{\centering\scriptsize{Integrated Gradients}}
      \fbox{\includegraphics[width=.95\linewidth]{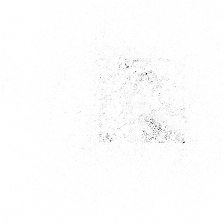}}
      \fbox{\includegraphics[width=.95\linewidth]{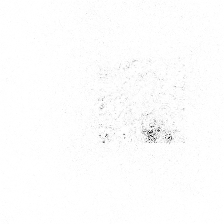}}
    \end{subfigure}%
    \begin{subfigure}[b]{.1\textwidth}
      \centering
      \caption{\centering\scriptsize{DeepSHAP}}
      \fbox{\includegraphics[width=.95\linewidth]{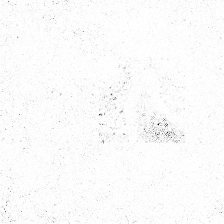}}
      \fbox{\includegraphics[width=.95\linewidth]{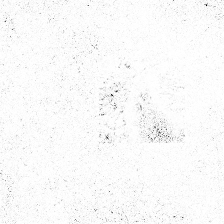}}
    \end{subfigure}%
    \begin{subfigure}[b]{.1\textwidth}
      \centering
      \caption{\centering\scriptsize{IBA}}
      \fbox{\includegraphics[width=.95\linewidth]{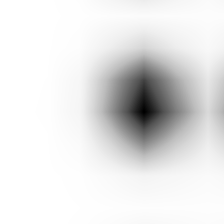}}
      \fbox{\includegraphics[width=.95\linewidth]{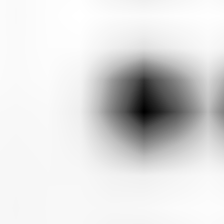}}
    \end{subfigure}%
    \begin{subfigure}[b]{.1\textwidth}
      \centering
      \caption{\centering\scriptsize{Extremal Perturbation}}
      \fbox{\includegraphics[width=.95\linewidth]{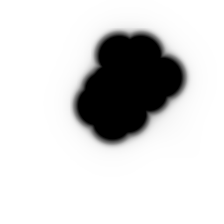}}
      \fbox{\includegraphics[width=.95\linewidth]{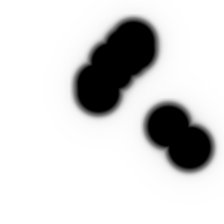}}
    \end{subfigure}%
  \caption{\textbf{Class Sensitivity - Single Feature Scenario}: The feature is generated such that it contributes to one output and is null for another output. The first output is presented in the first row. The output to which the feature is null is presented below. The explanations are presented for both outputs. It is expected that the explanations for the two outputs differ. Moreover, the explanations should not attribute to the feature for the null output (second row.). The only method that attributes correctly, in this case, is GradCAM.}
 \label{fig:single_patch}
\end{figure*}

In this setting, we evaluate class sensitivity in the case where only one contributing feature is available. Suppose the explanation for the output of the corresponding feature is similar to the explanation for an output to which the feature does not contribute. In that case, the explanation is not sensitive to the output.
A visual example for this case is provided in \cref{fig:single_patch}. The results for the associated metric are provided in \cref{tab:experiments}.
In the single case scenario, we find new insights regarding the explanations. In this scenario, more explanation methods are prone to attribute the output to the single feature within the image. Interestingly the only method that is sensitive to output, in this case, is GradCAM. The explanation focuses on features other than the generated patch when applied to the output to which the feature is null. Even IBA and Extremal perturbation that performed well in double feature scenario identify the same feature for the two outputs. This might be the property for all perturbation/removal-based methods, that they converge to the only predictive feature within the input, even if the feature is predictive for another class. We provide examples of this phenomenon for IBA in supplementary materials.

\subsection{Feature Saturation}
\begin{figure*}[ht]
\centering
    \begin{subfigure}[b]{.1\textwidth}
      \centering
      \caption{\centering\scriptsize{Image}}
      {\fbox{\includegraphics[width=.95\linewidth]{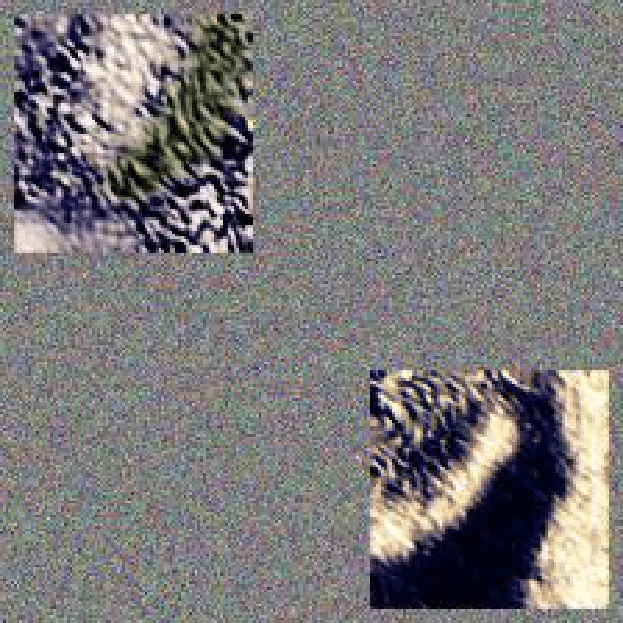}}}
      {\fbox{\includegraphics[width=.95\linewidth]{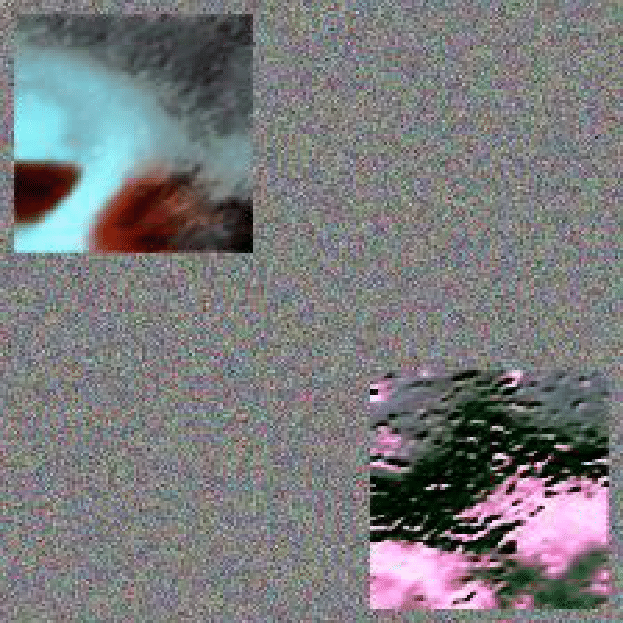}}}
    \end{subfigure}%
    \begin{subfigure}[b]{.1\textwidth}
      \centering
      \caption{\centering\scriptsize{GradCAM}}
      \fbox{\includegraphics[width=.95\linewidth]{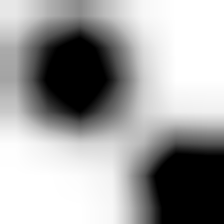}}
      \fbox{\includegraphics[width=.95\linewidth]{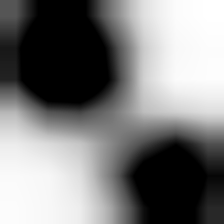}}
    \end{subfigure}%
    \begin{subfigure}[b]{.1\textwidth}
      \centering
      \caption{\centering\scriptsize{GradCAM++}}
      \fbox{\includegraphics[width=.95\linewidth]{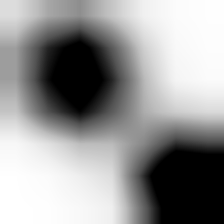}}
      \fbox{\includegraphics[width=.95\linewidth]{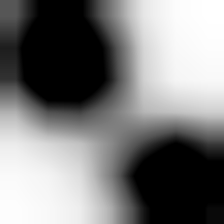}}
    \end{subfigure}%
    \begin{subfigure}[b]{.1\textwidth}
      \centering
      \caption{\centering\scriptsize{Gradient}}
      \fbox{\includegraphics[width=.95\linewidth]{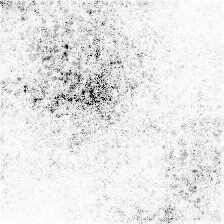}}
      \fbox{\includegraphics[width=.95\linewidth]{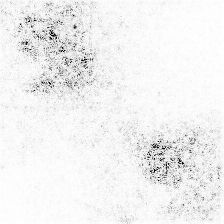}}
    \end{subfigure}%
    \begin{subfigure}[b]{.1\textwidth}
      \centering
      \caption{\centering\scriptsize{FullGrad}}
      \fbox{\includegraphics[width=.95\linewidth]{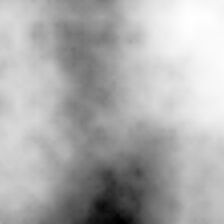}}
      \fbox{\includegraphics[width=.95\linewidth]{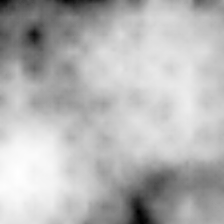}}
    \end{subfigure}%
    \begin{subfigure}[b]{.1\textwidth}
      \centering
      \caption{\centering\scriptsize{GuidedBackProp}}
      \fbox{\includegraphics[width=.95\linewidth]{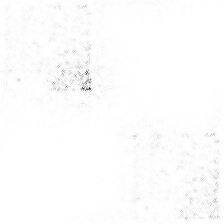}}
      \fbox{\includegraphics[width=.95\linewidth]{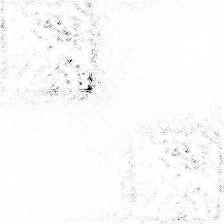}}
    \end{subfigure}%
    \begin{subfigure}[b]{.1\textwidth}
      \centering
      \caption{\centering\scriptsize{Integrated Gradients}}
      \fbox{\includegraphics[width=.95\linewidth]{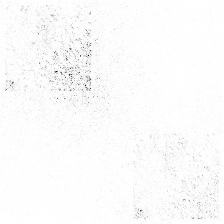}}
      \fbox{\includegraphics[width=.95\linewidth]{figures/Repeated/1/IntegratedGradients.png}}
    \end{subfigure}%
    \begin{subfigure}[b]{.1\textwidth}
      \centering
      \caption{\centering\scriptsize{DeepSHAP}}
      \fbox{\includegraphics[width=.95\linewidth]{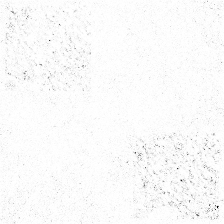}}
      \fbox{\includegraphics[width=.95\linewidth]{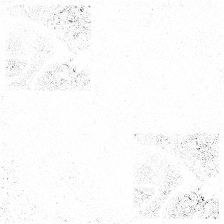}}
    \end{subfigure}%
    \begin{subfigure}[b]{.1\textwidth}
      \centering
      \caption{\centering\scriptsize{IBA}}
      \fbox{\includegraphics[width=.95\linewidth]{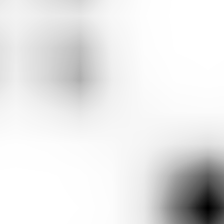}}
      \fbox{\includegraphics[width=.95\linewidth]{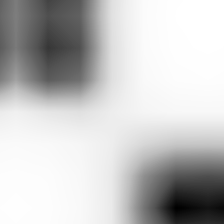}}
    \end{subfigure}%
    \begin{subfigure}[b]{.1\textwidth}
      \centering
      \caption{\centering\scriptsize{Extremal Perturbation}}
      \fbox{\includegraphics[width=.95\linewidth]{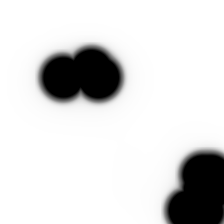}}
      \fbox{\includegraphics[width=.95\linewidth]{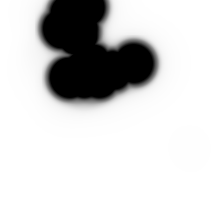}}
    \end{subfigure}%
  \caption{\textbf{Feature Saturation Experiment}: Each row is a sample from the feature saturation experiment. In this experiment, the features (patches) each saturate the output. In other terms, each individually generates the same output as their combination. A desired property for the attribution method is to distribute the contribution equally between the features. We observe that Extremal Perturbation and IBA can lean toward attributing the output to only one of the features. The formulation of these two method is based on keeping a region that keeps the output prediction. Thus, it is expected that they lean toward one feature.}
 \label{fig:feature_satutation_main}
\end{figure*}

This experiment aims to check how an attribution method behaves in case there are saturated features present in the input. The desirable property, in this case, is to attribute to both features. The results of the metric are provided in \cref{tab:experiments}. It is more intuitive to investigate the phenomenon in a visual example as presented in \cref{fig:feature_satutation_main}. The two features (patches) in the input contribute equally to the output, and the presence of only one is enough for the exact output prediction. We expect to observe that a method such as Extremal Perturbation attributing to only one of the features. The method searches for the smallest region that keeping it would keep the output prediction. In the case of saturated features, this translates to keeping only one feature. Note that it is also possible for the method to select part of each features. The metric in \cref{tab:experiments} shows that statistically, the method converges to one of the features. In the provided visual example, we also observe that the method is attributing to one of the features. The other methods are mostly attributing to both features, but considering the results from the Null player experiment and the Class Sensitivity experiment, the observation better be interpreted with caution. 
Several methods might be attributing to both features because they attribute to all features for other reasons. For instance, we observed with GuidedBackprop that the method is attributing to null feature and is attributing to both features when explaining different outputs. 
Thus, for this experiment, the prior knowledge that is derived from the previous ones changes the interpretation of the results. A method that is attributing to both features, but does that in null feature case as well, is not doing the attribution for fair distribution, but for other reasons.

\subsection{Limitations}
The optimization is challenging due to finding the right balance between the loss terms. We have used focal loss and hyperparameter tuning for this purpose.
One may also consider the required optimization time a challenge. Nevertheless, the framework has to be set up only once.
Coming up with more complex settings, such as adding more features, is challenging as the interactions between features add more optimization terms.

\section{Conclusion}
This work proposes an experimental framework for axiomatic evaluation of explanation methods using the model. Within the framework, the explanations are checked whether they comply with an axiom or satisfy a property. The experimental setup is realized through generating features using the model. Through feature generation, several scenarios for evaluating axioms are introduced. The framework reveals that many explanation methods identify a null feature as salient, even though the framework guarantees the feature to have no contribution.
Moreover, the framework shows many explanations are not class sensitive and generate roughly equivalent explanations for different outputs. The only methods that do not attribute to null features and are class sensitive are GradCAM, IBA, and Extremal Perturbations. We further analyze IBA and GradCAM++ on various layers of a neural network and reveal that the axioms are complied with only if they are applied to the final layer. We further demonstrate the property that IBA and Extremal perturbations do not attribute to all contributing features when features are saturated. Our current experiments in the proposed framework can be used to evaluate upcoming explanation methods. Furthermore, researchers can add more creative experiments to the proposed framework to assess explanations from other perspectives. 

\paragraph{Acknowledgement}
The work is supported by Munich Center for Machine Learning (MCML) with funding from the Bundesministerium fur Bildung und Forschung (BMBF) under the project 01IS18036B.

{\small
\bibliographystyle{ieee_fullname}
\bibliography{refs}
}

\clearpage
\section{Appendix}


\subsection{Null Feature Experiment}
\begin{figure*}[ht]
\centering
    \begin{subfigure}[b]{.1\textwidth}
      \centering
      \caption{\centering\scriptsize{Image}}
      {\fbox{\includegraphics[width=.95\linewidth]{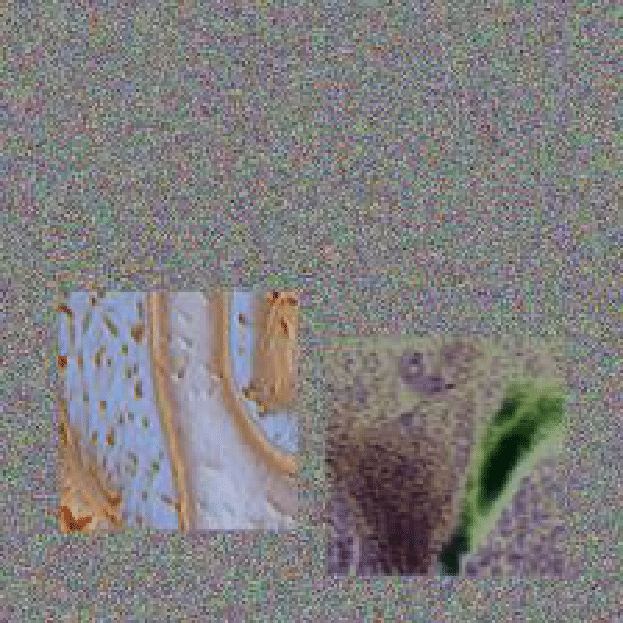}}}
      {\fbox{\includegraphics[width=.95\linewidth]{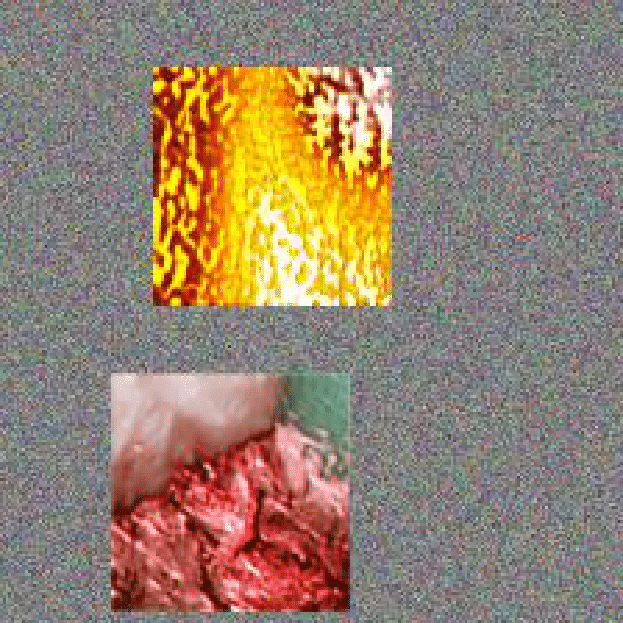}}}
      {\fbox{\includegraphics[width=.95\linewidth]{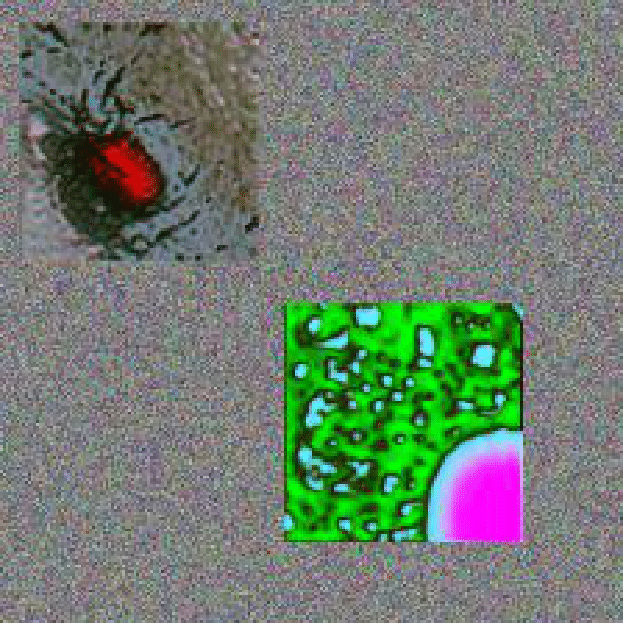}}}
      {\fbox{\includegraphics[width=.95\linewidth]{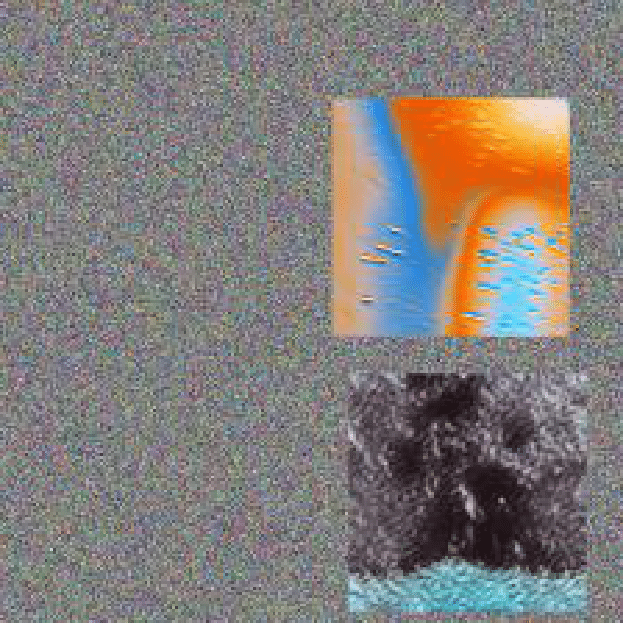}}}
      {\fbox{\includegraphics[width=.95\linewidth]{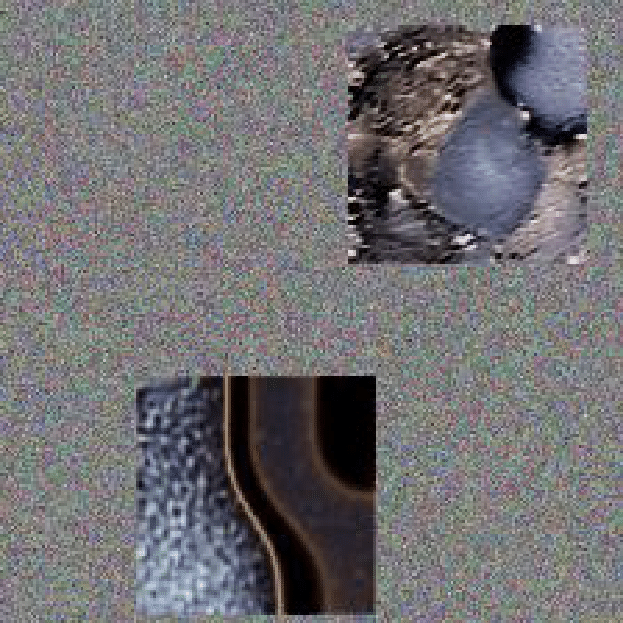}}}
    \end{subfigure}%
    \begin{subfigure}[b]{.1\textwidth}
      \centering
      \caption{\centering\scriptsize{GradCAM}}
      \fbox{\includegraphics[width=.95\linewidth]{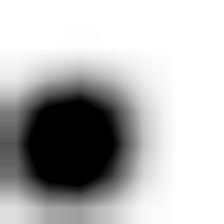}}
      \fbox{\includegraphics[width=.95\linewidth]{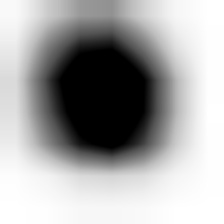}}
      {\fbox{\includegraphics[width=.95\linewidth]{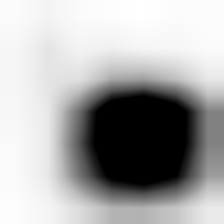}}}
      \fbox{\includegraphics[width=.95\linewidth]{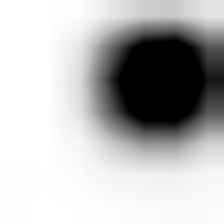}}
      \fbox{\includegraphics[width=.95\linewidth]{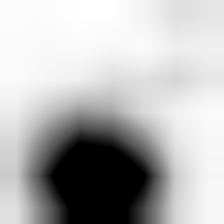}}
    \end{subfigure}%
    \begin{subfigure}[b]{.1\textwidth}
      \centering
      \caption{\centering\scriptsize{GradCAM++}}
      \fbox{\includegraphics[width=.95\linewidth]{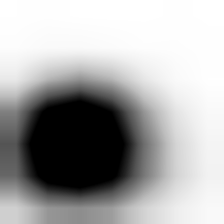}}
      \fbox{\includegraphics[width=.95\linewidth]{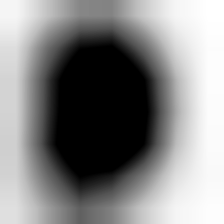}}
      \fbox{\includegraphics[width=.95\linewidth]{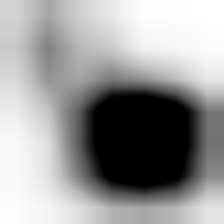}}
      \fbox{\includegraphics[width=.95\linewidth]{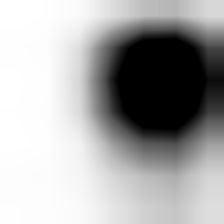}}
      \fbox{\includegraphics[width=.95\linewidth]{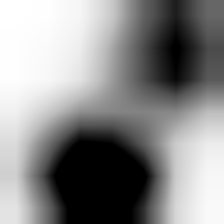}}
    \end{subfigure}%
    \begin{subfigure}[b]{.1\textwidth}
      \centering
      \caption{\centering\scriptsize{Gradient}}
      \fbox{\includegraphics[width=.95\linewidth]{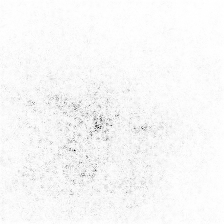}}
      \fbox{\includegraphics[width=.95\linewidth]{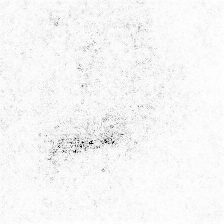}}
      \fbox{\includegraphics[width=.95\linewidth]{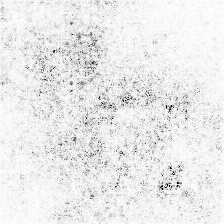}}
      \fbox{\includegraphics[width=.95\linewidth]{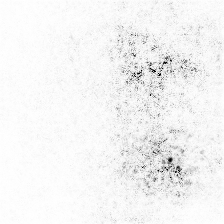}}
      \fbox{\includegraphics[width=.95\linewidth]{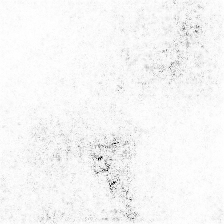}}
    \end{subfigure}%
    \begin{subfigure}[b]{.1\textwidth}
      \centering
      \caption{\centering\scriptsize{FullGrad}}
      \fbox{\includegraphics[width=.95\linewidth]{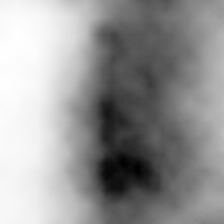}}
      \fbox{\includegraphics[width=.95\linewidth]{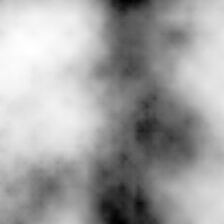}}
      \fbox{\includegraphics[width=.95\linewidth]{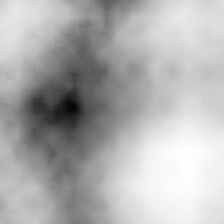}}
      \fbox{\includegraphics[width=.95\linewidth]{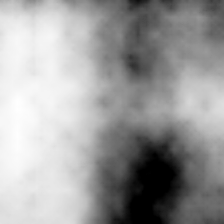}}
      \fbox{\includegraphics[width=.95\linewidth]{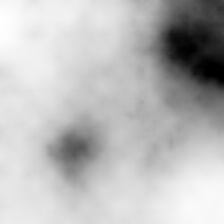}}
    \end{subfigure}%
    \begin{subfigure}[b]{.1\textwidth}
      \centering
      \caption{\centering\scriptsize{GuidedBackProp}}
      \fbox{\includegraphics[width=.95\linewidth]{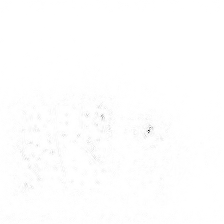}}
      \fbox{\includegraphics[width=.95\linewidth]{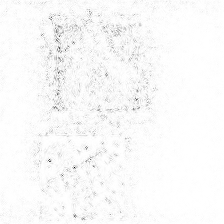}}
      \fbox{\includegraphics[width=.95\linewidth]{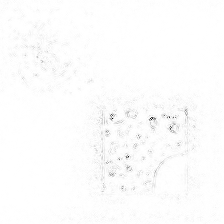}}
      \fbox{\includegraphics[width=.95\linewidth]{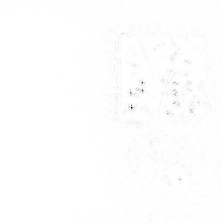}}
      \fbox{\includegraphics[width=.95\linewidth]{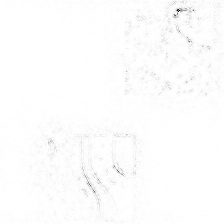}}
    \end{subfigure}%
    \begin{subfigure}[b]{.1\textwidth}
      \centering
      \caption{\centering\scriptsize{Integrated Gradients}}
      \fbox{\includegraphics[width=.95\linewidth]{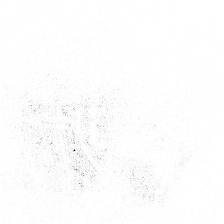}}
      \fbox{\includegraphics[width=.95\linewidth]{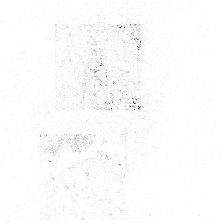}}
      \fbox{\includegraphics[width=.95\linewidth]{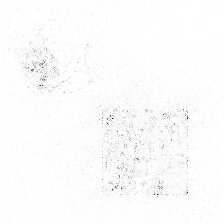}}
      \fbox{\includegraphics[width=.95\linewidth]{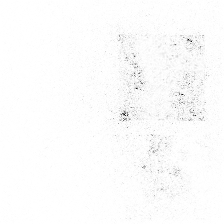}}
      \fbox{\includegraphics[width=.95\linewidth]{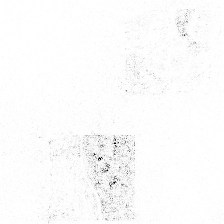}}
    \end{subfigure}%
    \begin{subfigure}[b]{.1\textwidth}
      \centering
      \caption{\centering\scriptsize{DeepSHAP}}
      \fbox{\includegraphics[width=.95\linewidth]{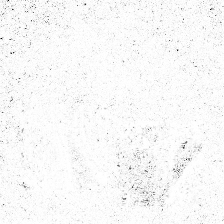}}
      \fbox{\includegraphics[width=.95\linewidth]{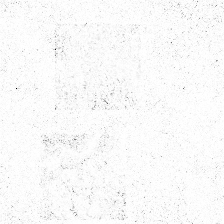}}
      \fbox{\includegraphics[width=.95\linewidth]{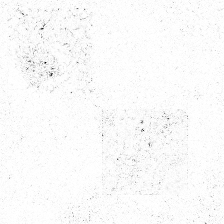}}
      \fbox{\includegraphics[width=.95\linewidth]{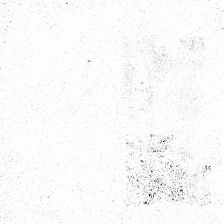}}
      \fbox{\includegraphics[width=.95\linewidth]{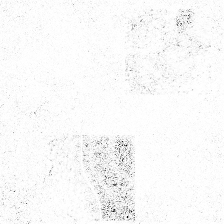}}
    \end{subfigure}%
    \begin{subfigure}[b]{.1\textwidth}
      \centering
      \caption{\centering\scriptsize{IBA}}
      \fbox{\includegraphics[width=.95\linewidth]{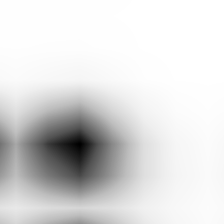}}
      \fbox{\includegraphics[width=.95\linewidth]{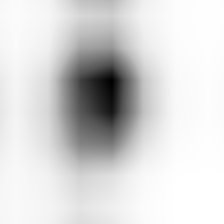}}
      \fbox{\includegraphics[width=.95\linewidth]{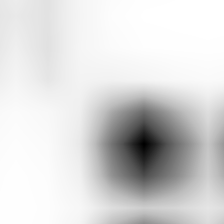}}
      \fbox{\includegraphics[width=.95\linewidth]{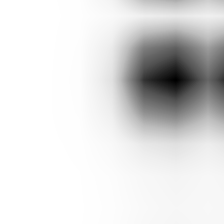}}
      \fbox{\includegraphics[width=.95\linewidth]{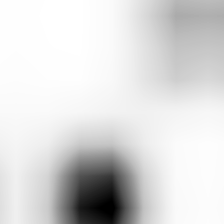}}
    \end{subfigure}%
    \begin{subfigure}[b]{.1\textwidth}
      \centering
      \caption{\centering\scriptsize{Extremal Perturbation}}
      \fbox{\includegraphics[width=.95\linewidth]{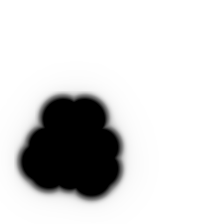}}
      \fbox{\includegraphics[width=.95\linewidth]{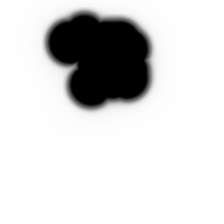}}
      \fbox{\includegraphics[width=.95\linewidth]{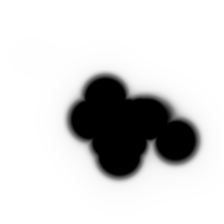}}
      \fbox{\includegraphics[width=.95\linewidth]{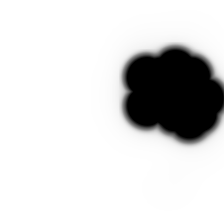}}
      \fbox{\includegraphics[width=.95\linewidth]{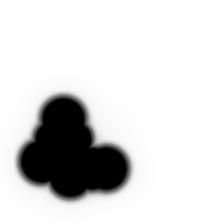}}
    \end{subfigure}%
  \caption{\textbf{Null Feature Experiment}: Each row represents a sample from the null feature experiment. In each row, the image on the left represents the generated features on the reference (noise) input. The features are generated using the model itself. Within the image, the lower feature (patch) is generated such that it is a null feature for the output. The rest of the images represent different explanations. As the second feature is a null feature, an explanation method should not assign importance to it. We observe that GradCAM, IBA, and Extremal Perturbation perform best in avoiding the null feature.}
 \label{fig:null_player_supplementary}
\end{figure*}

More visual examples for the null feature experiment are provided in \cref{fig:null_player_supplementary}.


\subsection{Optimization (Focal Loss) Details}
The following terms each indicate the keys used during optimization phase as focal loss entries.

\begin{equation}
    \begin{aligned}
    \kappa_{\mathcal{L}} &= \sigma_t(\Phi_t(X))\\
    \kappa_{\{f_a\}} &= tanh(\frac{|\Phi_t(X_{\{f_a\}}) - \Phi_t(X_{\{\}})|}{\min (|\Phi_t(X_{\{f_a\}})|,\,|\Phi_t(X_{\{\}})|)})\\
    \kappa_{\{f_a,\,f_b\}} &= tanh(\frac{|\Phi_t(X_{\{f_a,\,f_b\}}) - \Phi_t(X_{\{f_a\}})|}{\min (|\Phi_t(X_{\{f_a,\,f_b\}})|,\,|\Phi_t(X_{\{f_a\}})|)})
    \end{aligned}
\end{equation}

where $\sigma_t(.)$ designates softmax function corresponding to the target $t$. The weighted average of the term $\kappa_\mathcal{L}$ is emploied in each of the loss terms as the multiplicand of the first term. The weighted average of other two are used in their corresponding scenarios as the multipicand of the latter terms. The weighted average through each iteration is calculated in the following manner:

\begin{equation}
    \begin{aligned}
     \hat{\kappa}_{t+1} =  \alpha \kappa_t + (1-\alpha) \hat{\kappa}_t 
    \end{aligned}
\end{equation}
 During all the experiments, the $\alpha$ is set to $0.1$, and the intitial value for $\kappa$ is $0.5$. To further faciliate the optimization phase, we initially optmize the features to maximize $\Phi_t(X)$ fo their designated target class.

\end{document}